%% file: 44422_supp.tex
\documentclass[10pt,twocolumn,letterpaper]{article}

\usepackage{cvpr}              
\usepackage{array}
\usepackage{textcomp}
\usepackage{stfloats}
\usepackage{url}
\usepackage{verbatim}
\usepackage{graphicx}
\usepackage{xcolor}        
\usepackage{multirow} 
\usepackage{multicol} 
\usepackage{lineno}
\usepackage{bm}
\usepackage{threeparttable}
\usepackage{pifont}
\usepackage[table]{xcolor}
\usepackage{makecell}
\usepackage{algorithm}
\usepackage{algorithmic}
\usepackage{color}

\usepackage{array}
\usepackage{cuted}
\usepackage{capt-of}

\usepackage{caption}
\definecolor{WhiteSmoke}{rgb}{1.0000 1.0000	0.8784}
\definecolor{LightBlue}{rgb}{0.9098 0.9686 0.8941}
\definecolor{Ivory}{rgb}{1, 1, 0.9412}
\definecolor{LightGoldenrodYellow}{rgb}{0.9804	0.9804	0.8235}
\definecolor{LightCyan}{rgb}{0.8784 1 1}

\input{preamble}
\definecolor{cvprblue}{rgb}{0.21,0.49,0.74}
\usepackage[pagebackref,breaklinks,colorlinks,allcolors=cvprblue]{hyperref}

\def\paperID{44422} 
\def\confName{CVPR}
\def\confYear{2026}

\title{Bilevel Layer-Positioning LoRA for Real Image Dehazing}

\author{Yan Zhang$^{1}$
	\and Long Ma$^{3}$
	\and Yuxin Feng$^{4}$ 
	\and Zhe Huang$^{1}$
	\and Fan Zhou$^{1,2}$
	\and Zhuo Su$^{1,2,}\thanks{Corresponding author (suzhuo3@mail.sysu.edu.cn).}$
	\and  \normalsize $^{1}$School of Computer Science and Engineering, Sun Yat-sen University\\
	\normalsize$^2$National Engineering Research Center of Digital Life$\;\;$
	\normalsize$^3$Dalian University of Technology$\;\;$
	\normalsize$^4$Xidian University
}

\begin{document}
\maketitle
\input{sec/0_abstract}    
\input{sec/1_intro}

\input{sec/2_formatting}

{
    \small
    \bibliographystyle{ieeenat_fullname}
    \bibliography{main}
}


\end{document}

%% file: sec/0_abstract.tex
\thispagestyle{empty}
\begin{strip}
	\vspace{-1.47cm}
	\centering
	\begin{tabular}{c} 
	\includegraphics[width=0.98\textwidth]{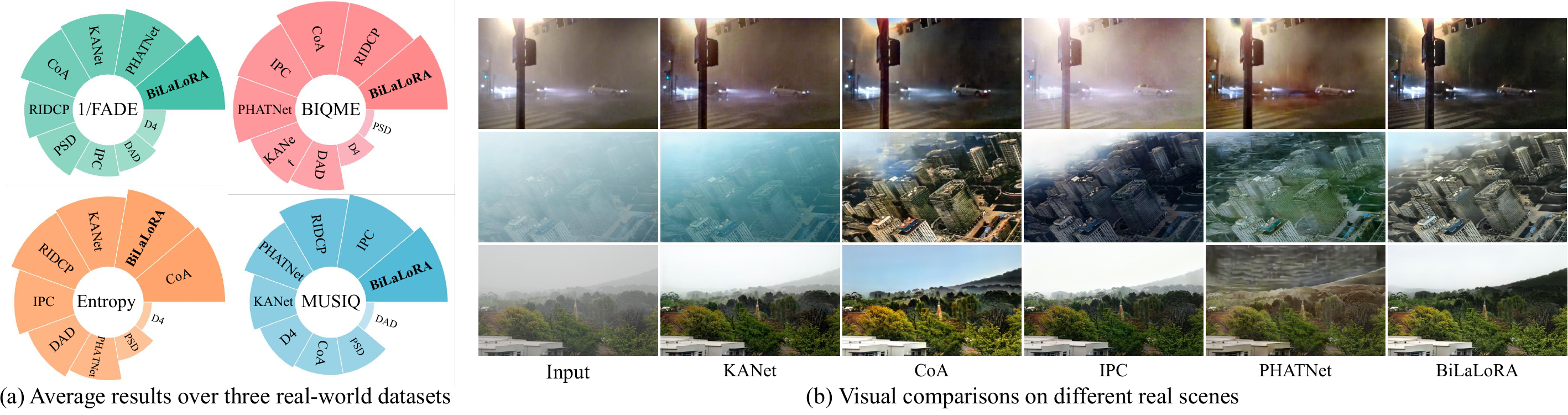}\\
	\end{tabular}
	\vspace{-0.3cm}
	\captionof{figure}{\textbf{Performance comparison}. Sub-figure (a) shows quantitative results on four non-reference metrics across three real-world datasets (RTTS~\cite{reside}, URHI~\cite{reside}, and Fattal~\cite{fattal}), while sub-figure (b) presents visual comparisons on different challenging real scenes.}
	\label{fig:FirstFigure}
\end{strip}

\begin{abstract}
Learning-based real image dehazing methods have achieved notable progress, yet they still face adaptation challenges in diverse real haze scenes. These challenges mainly stem from the lack of effective unsupervised mechanisms for unlabeled data and the heavy cost of full model fine-tuning. To address these challenges, we propose the haze-to-clear text-directed loss that leverages CLIP’s cross-modal capabilities to reformulate real image dehazing as a semantic alignment problem in latent space, thereby providing explicit unsupervised cross-modal guidance in the absence of reference images. Furthermore, we introduce the {Bi}level {La}yer-positioning {LoRA} (\textbf{BiLaLoRA}) strategy, which learns both the LoRA parameters and automatically search the injection layers, enabling targeted adaptation of critical network layers. Extensive experiments demonstrate our superiority against state-of-the-art methods on multiple real-world dehazing benchmarks. The code is publicly available at {\color{blue} https://github.com/YanZhang-zy/BiLaLoRA}.

\end{abstract}
\vspace{-0.3cm}

%% file: sec/1_intro.tex
\section{Introduction}
\label{sec:intro}
Image dehazing, as one of the core research directions in low-level vision tasks, aims to remove atmospheric degradation caused by scattering and particulate absorption, enhancing visual quality and facilitating downstream tasks.


Traditional dehazing algorithms primarily rely on handcrafted priors, yet idealized assumptions limit their applicability~\cite{DCP}. In recent years, deep learning-based methods have demonstrated remarkable performance on synthetic datasets~\cite{11306247}~\cite{10947620}~\cite{ju2025towards}~\cite{sci++}. These methods typically synthesize haze images based on the atmospheric scattering model, but the modeling process often oversimplifies the complex light field distribution and medium characteristics in the atmosphere, leading to significant domain gap between synthetic data and real-world scenarios~\cite{reside}~\cite{haze4k}. To mitigate this issue, researchers have begun exploring domain adaptation strategies to improve generalization capability~\cite{FENG2026113301}~\cite{DAD}. Despite notable progress, existing approaches still face two critical challenges:
\begin{itemize}
	\item \textbf{Lack of effective unsupervised mechanisms}: Clean ground truth is hard to acquire in real scenes, so models often rely on synthetic data or weak priors, making \textit{it crucial to design effective unsupervised objectives for unpaired haze images to achieve robust dehazing across diverse real-world conditions.}
	\item \textbf{The heavy cost of full model fine-tuning}: Even with effective unsupervised objectives, real image dehazing often depends on updating all network parameters, leading to high computational and memory costs. \textit{This hinders fast adaptation in practical deployments, making it necessary to develop more efficient adaptation strategies.}
	
\end{itemize}
\begin{figure*}[!t]
	\centering
	\includegraphics[width=0.98\textwidth]{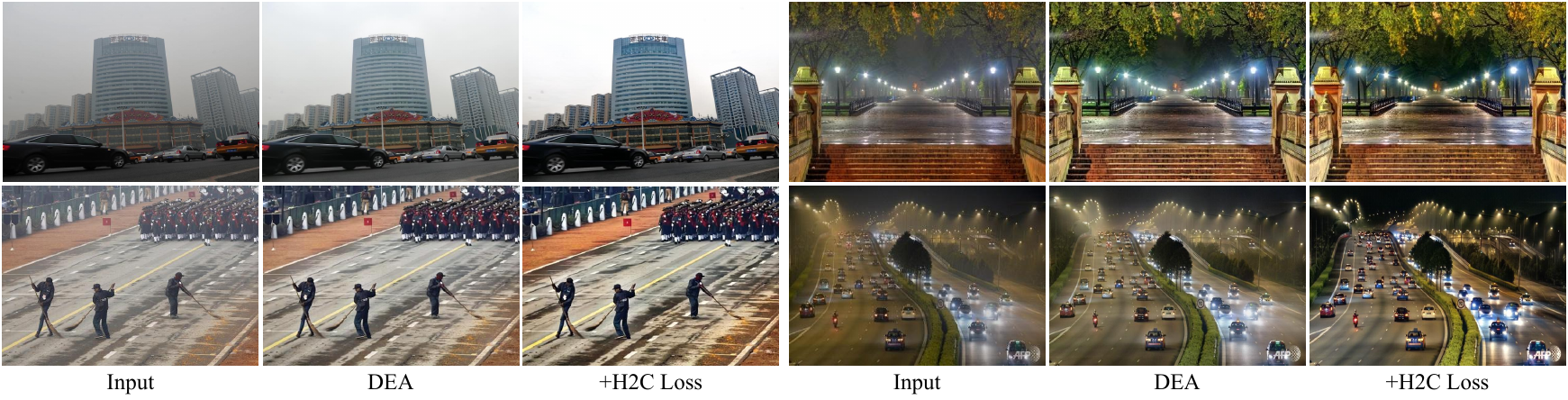}
	\vspace{-0.2cm}
	\caption{\textbf{Effectiveness of H2C loss in different real-scenes. } The left two rows (daytime scenes) are from URHI and Fattal, and the right two rows (nighttime scenes) are from NHRW~\cite{NHRW}.}
	\label{fig:H2CEffects}
	\vspace{-1em}
\end{figure*}
\subsection{Our Main Contributions}
To address the domain adaptation challenges faced by dehazing models trained on synthetic data when applied to real-world scenes, we propose the Bilevel Layer-positioning LoRA (BiLaLoRA) framework. As shown in Fig.~\ref{fig:FirstFigure}, BiLaLoRA outperforms existing methods across three real-world datasets. Specifically, we first utilize the CLIP model~\cite{CLIP} to design a Haze-to-Clear text-directed loss (H2C), reframing the dehazing task as a semantic mapping from haze to clear images. As shown in Fig.~\ref{fig:H2CEffects}, the H2C loss significantly enhances the performance of pre-trained models across various real-world scenes. However, Table~\ref{tab:comparison} shows that applying the H2C loss with full fine-tuning incurs high computational costs with limited adaptability.

Through comprehensive analysis of the domain adaptation process, we observe that the performance bottleneck layers affected by the domain gap vary dynamically depending on the model architecture and scene characteristics. To address this dynamic behavior,  BiLaLoRA employs low-rank adaptation for parameter-efficient fine-tuning and unifies the selection of adapter injection layers with weight optimization into a bilevel optimization problem, enabling more precise and efficient cross-domain adaptation. The main contributions are summarized as follows:

\begin{itemize}
	\item To address the lack of effective unsupervised mechanisms, we propose the H2C loss by leveraging CLIP's cross-modal capability. It reformulates the dehazing process as a semantic alignment task in latent space, enabling flexible optimization without requiring paired real data.

	\item To mitigate the heavy cost of full model fine-tuning, we introduce BiLaLoRA, an efficient adaptation strategy. Through bi-level optimization of both the LoRA  positions and weights, it automatically pinpoints and fine-tunes bottleneck layers without manual configuration.

	\item With minimal computational and storage overhead, BiLaLoRA achieves efficient transfer from synthetic to real domains. Its plug-and-play nature supports rapid switching multiple target domains, achieving an optimal balance among performance, efficiency, and flexibility.
\end{itemize}

\section{Related Works}

\subsection{Real Image Dehazing}
Real image dehazing aims to mitigate the negative effects of natural haze on visual quality as well as restore the colors and details. Although deep learning-based dehazing models demonstrate strong performance on synthetic datasets, they exhibit significant performance degradation in real-world scenes due to the synthetic-to-real domain gap. To bridge this gap, researchers have developed various strategies to reduce the distribution discrepancy between domains. Domain adaptation methods enhance cross-domain generalization by aligning feature distributions between source and target domains~\cite{Breaking}. Physics-based prior-constrained methods integrate the atmospheric scattering model into the network design, guiding the model to learn more physically interpretable representations~\cite{PSD}~\cite{CORUN}. Unsupervised approaches leverage Generative Adversarial Networks (GANs) to learn the dehazing mapping from unpaired data, thereby alleviating the dependency on large-scale labeled datasets~\cite{D4}~\cite{UCL}. Other solutions involve using test-time adaptation for dynamically adjusting feature statistics~\cite{PTTD} and leveraging pre-trained visual priors like VQGAN to boost content generation~\cite{RIDCP}~\cite{IPC}. Nevertheless, these approaches are typically constrained by complex training pipelines and substantial computational overhead, while lacking the flexibility necessary to accommodate diverse real-world degradation patterns.

\subsection{Parameter-Efficient Fine-Tuning}
Parameter-Efficient Fine-Tuning (PEFT) offers a solution to the aforementioned challenges. The core principle of PEFT is to freeze the pre-trained model while adapting only a small number of additional parameters, thereby enabling rapid adaptation to new tasks at a minimal computational cost. Low-Rank Adaptation (LoRA)~\cite{LoRA} represents a seminal PEFT technique that achieves performance comparable to full fine-tuning by injecting trainable low-rank matrices into selected weight layers. Originally developed for natural language processing, LoRA has been successfully extended to computer vision tasks~\cite{lora_IR}. For instance, Sun et al.~\cite{sun2025pixel} decoupled pixel-level reconstruction from semantic enhancement through LoRA modules, achieving high-quality controllable image super-resolution. Similarly, Zhong et al.~\cite{LoRA_SAM} integrate LoRA with lightweight convolutions to incorporate local visual priors into the Segment Anything Model (SAM), substantially improving its cross-domain generalization in semantic segmentation. These successes demonstrate that LoRA provides an efficient and flexible paradigm for domain adaptation in vision tasks, particularly in resource constrained scenarios or situations requiring rapid adaptation to multiple target domains.

%% file: sec/2_formatting.tex
\section{The H2C Unsupervised Loss}\label{sec:loss}

\subsection{Latent Cross-Modality Consistency}
In real image dehazing, a fundamental challenge arises from the absence of paired ground-truth clear images. This inherent limitation renders supervised paradigms that rely on reference images and pixel-level loss functions inapplicable. However, humans distinguish haze and clear images not relying on pixel-wise correspondences, but through high-level semantic understanding of the scene. This observation suggests that if such semantic comprehension can be formulated as computable guidance signals, it may be possible to circumvent the dependency on reference images.

Fortunately, the advent of vision-language models like CLIP has provided new pathways for addressing this challenge~\cite{CLIPLIT}.  By constructing a unified latent embedding space, CLIP can effectively quantify the semantic content of images. Leveraging this capability, we propose a novel strategy based on the principle of cross-modality consistency, reconceptualizing the dehazing process as a semantic mapping from haze to clear and accordingly formulating an optimization objective that requires no paired supervision.
\begin{figure}[t]
	\centering
	\includegraphics[width=0.48\textwidth]{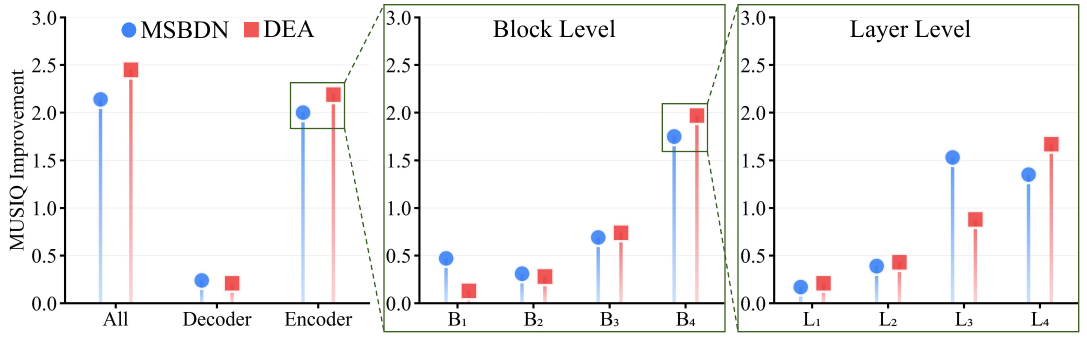}
	\vspace{-0.5cm}
	\caption{\textbf{Contribution of different network components to domain adaptation}.}
	\label{fig:Motivation}
		\vspace{-1.3em}
\end{figure}

\subsection{H2C: Haze-to-Clear Text-Directed Loss}
To translate high-level semantic information into quantifiable guidance, we design a haze-to-clear text-directed loss (H2C) to align the semantic transformation of dehazing process with a target direction defined by text prompts. Specifically, for the input $I_{\text{in}}$ and dehazing output $I_{\text{out}}$, we utilize the pre-trained CLIP image encoder to extract their feature $V_{\text{in}}$ and $V_{\text{out}}$, which serve as their coordinates in the semantic space. Meanwhile, we use the CLIP text encoder to extract features from negative text prompt $T_{\text{neg}}$ (a photo with haze) and positive text prompt $T_{\text{pos}}$ (a clear photo), thereby defining the start and end points of the ideal transformation.

Within the latent embedding space, the semantic transformation produced by the dehazing model is characterized by the displacement vector $\Delta V_{\text{img}} = V_{\text{out}} - V_{\text{in}}$, which encodes the perceptual shift from haze to clear image. Concurrently, we define a target direction vector $\Delta T_{\text{text}} = T_{\text{pos}} - T_{\text{neg}}$ that represents the desired semantic trajectory in the text-guided feature space. The central objective of the H2C loss is to promote directional alignment between the image semantic transformation $\Delta V_{\text{img}}$ and the text-guided target direction $\Delta T_{\text{text}}$. To quantify this alignment, we employ cosine similarity as follows:

\begin{algorithm}[t]
	\caption{Bilevel Layer-positioning LoRA}
	\label{alg:lora}
	\footnotesize
	\begin{algorithmic}[1]
		\REQUIRE Initial LoRA weights $\bm{\omega}_0$, architecture parameters $\bm{\alpha}_0$, 
		frozen backbone parameters $\bm{\theta}$, step-size $\eta_w$, $\eta_{\alpha}$, 
		total epochs $T$, switch epoch $T_s$, number of layers to select $k$.
		\ENSURE Optimized parameters $\bm{\omega}^*$, $\bm{\alpha}^*$.
		\STATE \% \textit{The Bilevel Layer-Positioning Stage}
		\FOR{$t = 0 $ : $T_{s}-1 $}
		\STATE \% \textit{Update the architecture parameters}
		\STATE {Calculate gradients $\bm{g}_\alpha$ by Eq. \eqref{eq:7}}.
		\STATE $\bm{\alpha}_{t+1} = \bm{\alpha}_{t} - \eta_{\alpha} \bm{g}_\alpha(\bm{\omega}_{t}, \bm{\alpha}_{t}, \bm{\theta})$.	  		
		\STATE \% \textit{Update the LoRA weights}
		\STATE $\bm{\omega}_{t+1} = \bm{\omega}_{t} - \eta_w \nabla_{\bm{\omega}} f(\bm{\omega}_{t}, \bm{\alpha}_{t+1}, \bm{\theta})$.
		\ENDFOR
		\STATE \% \textit{Select top-k layers based on ranking scores}
		\STATE $\bm{\alpha}^* = \text{TopK}(\bm{\alpha}_{T_s}, k)$.
		\STATE \% \textit{The LoRA Fine-Tuning Stage}		
		\FOR{$t = T_{s}$ : $T$}
		\STATE \% \textit{Update the LoRA weights under the optimized $\bm{\alpha}^*$}
		\STATE $\bm{\omega}_{t+1} = \omega_{t} - \eta_w \nabla_{\bm{\omega}} f(\bm{\omega}_{t}, \bm{\alpha}^*, \bm{\theta})$.
		\ENDFOR		
		\STATE $\bm{\omega}^*=\bm{\omega}_{T}$.
		\STATE \textbf{return} $\bm{\omega}^*$, $\bm{\alpha}^*$.
	\end{algorithmic}
\end{algorithm}  
\begin{equation}\label{eq:1}
	L_\text{H2C}=1-{\frac{\Delta V_\text{img}\cdot\Delta T_\text{text}}{\|\Delta V_\text{img}\|_{2}\cdot|\Delta T_\text{text}\|_{2}}}.
\end{equation}

\subsection{Discussion}
Notably, unlike domain adaptation methods that constrain feature space distributions, the H2C loss guides the model to generate clearer images by leveraging semantic information in the latent embedding space, without introducing additional complex structures. Moreover, the H2C loss demonstrates strong generalization across multiple real-world scenes with significant domain differences by adjusting text prompts. As shown in Fig.~\ref{fig:H2CEffects}, DEA~\cite{DEA} effectively transfers from synthetic domain to real daytime scenes. Additionally, it adapts to nighttime scenes by using the prompt `a photo with nighttime haze'. By providing universal cross-domain guidance, the H2C loss effectively handles diverse haze patterns.

\section{BiLaLoRA}\label{sec:BiLaLoRA}
\subsection{Motivation}
To pinpoint the network components most critical for domain adaptation, we fine-tuned MSBDN~\cite{MSBDN} and DEA on real-domain data using the H2C loss, then quantified each component’s contribution by grafting adapted modules back into the original models and measuring the MUSIQ~\cite{musiq} improvements on the RTTS~\cite{reside} dataset. As shown in Fig.~\ref{fig:Motivation}, the encoder plays a crucial role, with its final block accounting for the majority of the performance gain. However, the specific layers within that block that drive the improvement differ substantially across architectures. This disparity indicates that the performance bottleneck caused by the domain gap is not static but shifts dynamically depending on the characteristics of the network architecture.
\begin{figure*}[t]
	\centering
	\includegraphics[width=0.98\textwidth]{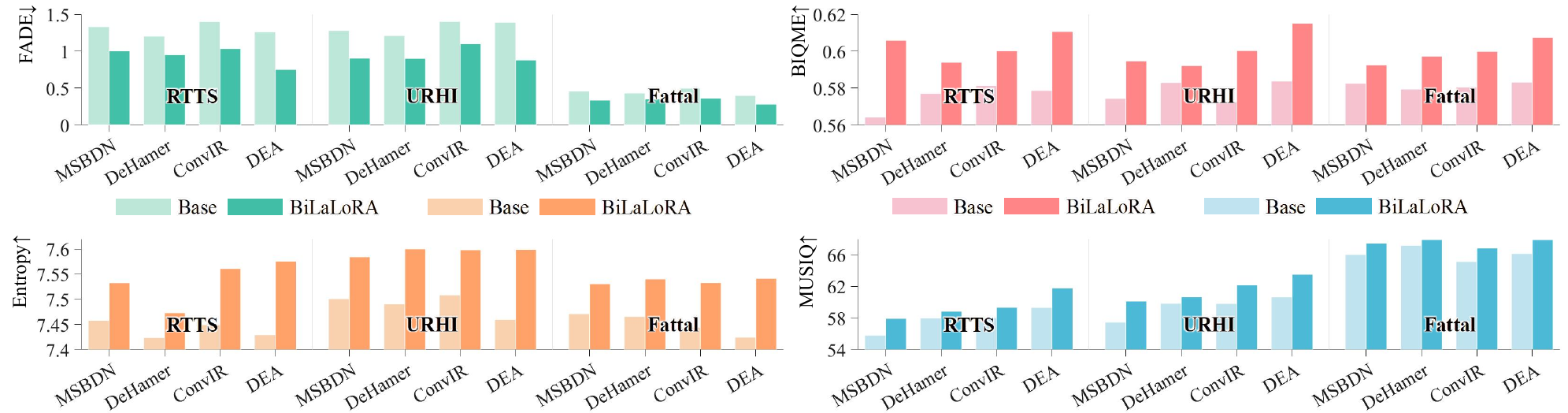}
	\vspace{-0.25cm}
	\caption{\textbf{Quantitative results of cross-model flexibility}. We evaluate four baseline dehazing architectures on three real dehazing datasets with four non-reference metrics to ensure the generality of this property. }
	\label{fig:flexibilityquan}
	\vspace{-0.5em}
\end{figure*}

\begin{figure}[t]
	\centering
	\includegraphics[width=0.463\textwidth]{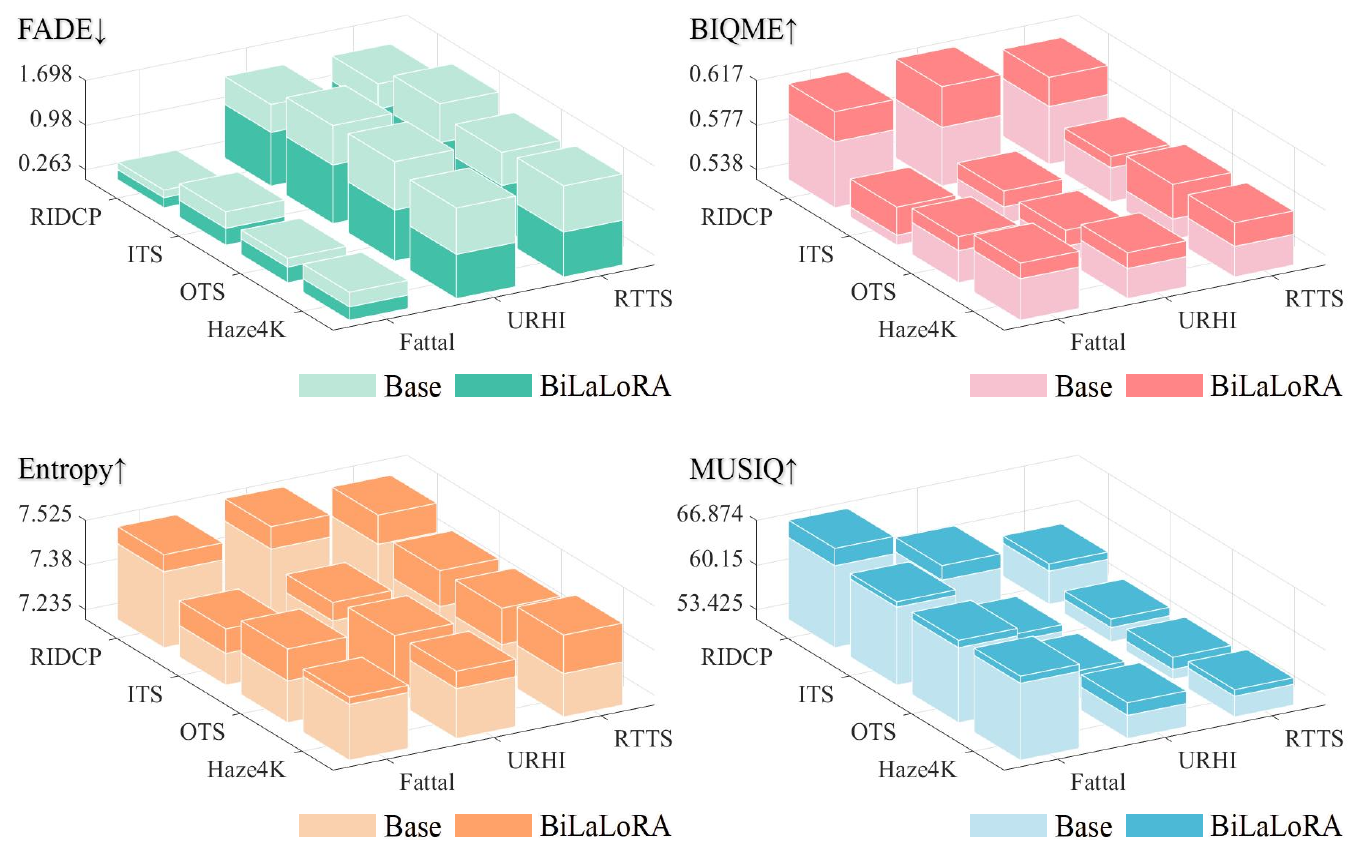}
	\vspace{-0.25cm}
	\caption{\textbf{Quantitative results of cross-domain stability}. We leverage DEA pre-trained on four synthetic datasets to verify robustness across different source domains.}
	\label{fig:stabilityquan}
			\vspace{-1em}
\end{figure}

\subsection{Model-Agnostic Layer-Positioning Modeling}
Building upon these observations, we identify the central challenge in efficient domain adaptation as the precise positioning and optimization of performance bottleneck layers. As PEFT approach, LoRA provides an ideal solution to this problem. For a pre-trained weight matrix $W_0 \in \mathbb{R}^{d_{\text{out}} \times d_{\text{in}}}$, LoRA parameterizes the weight update $\Delta W$ via a low-rank decomposition using trainable matrices $A \in \mathbb{R}^{r \times d_{\text{in}}}$ and $B \in \mathbb{R}^{d_{\text{out}} \times r}$, where $r \ll \min(d_{\text{out}}, d_{\text{in}})$ denotes the rank.

However, the effectiveness of LoRA is critically contingent upon the strategic selection of injection layers. Conventional approaches predominantly rely on heuristic or empirically driven choices, which lack generalizability across diverse architectural paradigms. To overcome this limitation, we reformulate the layer selection problem as a differentiable architecture search task. Specifically, we regulate each LoRA module with a learnable gating parameter $\alpha$, which is constrained to the range $(0,1)$ using a sigmoid function and modulates the contribution of the low-rank increment in conjunction with the scaling factor $\gamma$:

\begin{equation}\label{eq:2}
	W' = W_0 + \alpha \cdot \gamma \cdot \Delta W,
\end{equation}
in which $\alpha$ serves as a continuous relaxation of the discrete layer selection decision.

\begin{figure}[t]
	\centering
	\includegraphics[width=0.463\textwidth]{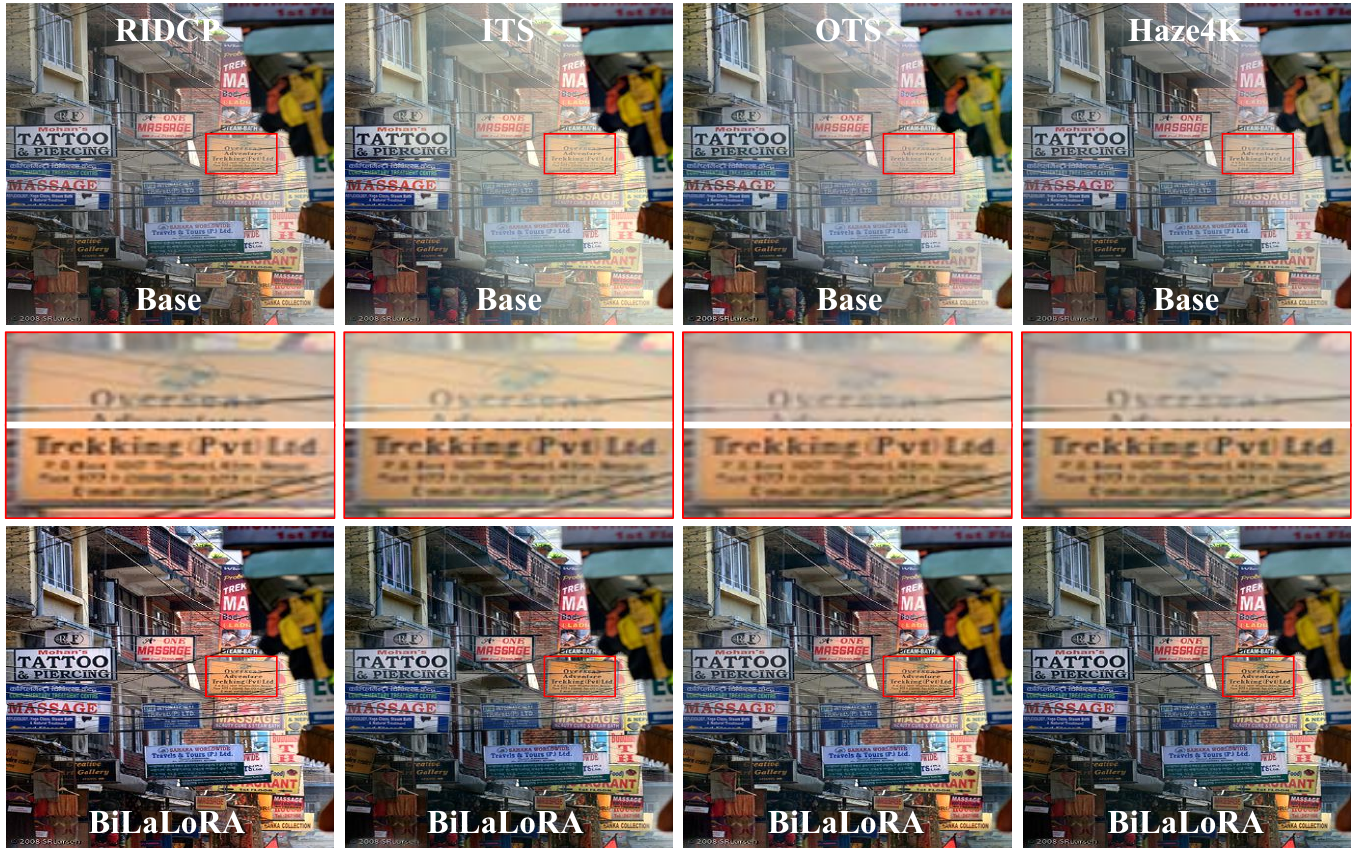}
	\vspace{-0.25cm}
	\caption{\textbf{Qualitative results of cross-domain stability}. BiLaLoRA consistently enhances real-world performance across various source domains, with data from Fattal.}
	\label{fig:stabilityqual}
		\vspace{-1em}
\end{figure}

\begin{figure*}[t]
	\centering
	\includegraphics[width=0.97\textwidth]{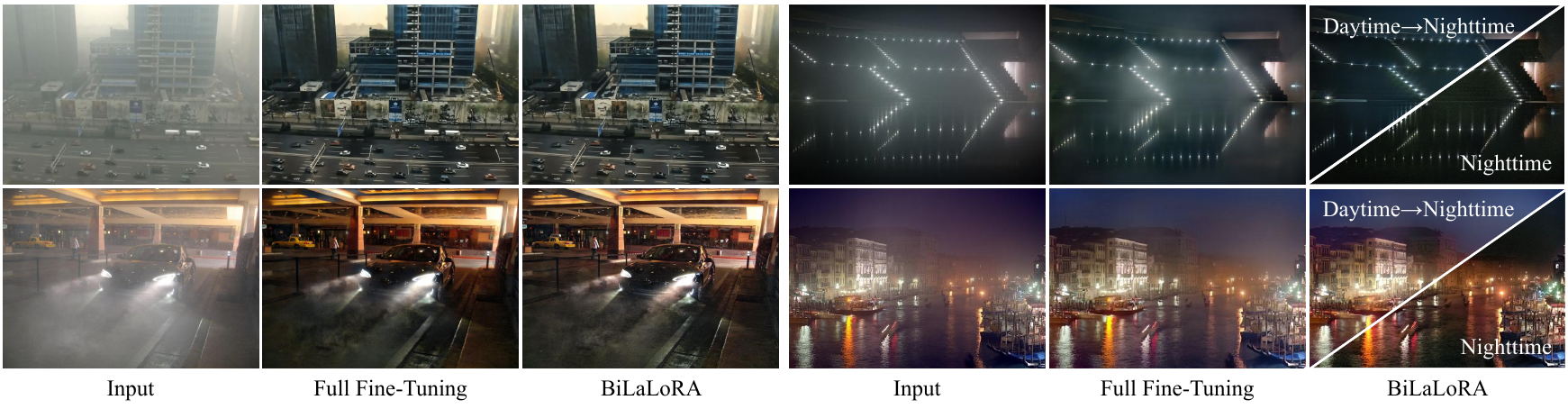}
	\vspace{-0.3cm}
	\caption{\textbf{Qualitative results of full fine-tuning vs. BiLaLoRA}. The left two rows are from RTTS and the right two rows are from NHRW.}
	\label{fig:LoRA}
	\vspace{-1em}
\end{figure*}

\subsection{BiLaLoRA: Bilevel Layer-Positioning LoRA}
Given the hierarchical dependency between architectural parameters and LoRA weight matrices, single-level optimization frameworks fail to capture this inherent relationship. To address this limitation, we propose Bilevel Layer-Positioning LoRA (BiLaLoRA), which formulates the learning objective as a bilevel optimization problem:
\begin{equation}\label{eq:3}
	\begin{aligned}
		& \min_{\bm{\alpha}}\, \varphi({\bm{\omega}^*}(\bm{\alpha}), \bm{\alpha}), \\
		& \text{s.t.}\,  \bm{\omega}^*(\bm{\alpha}) \in \arg\min_{\bm{\omega}} \, \psi(\bm{\omega}, \bm{\alpha}),
	\end{aligned}
\end{equation}
where $\varphi$ and $\psi$ denote BiLaLoRA’s upper and lower objectives for optimizing the architectural parameters governing the layer selection $\bm{\alpha}$ and low-rank weight increments $\Delta W$, respectively.  The principal computational challenge lies in evaluating the hypergradient $\nabla_{\bm{\alpha}} \varphi$ for the upper-level objective. Due to the implicit dependence of $\bm{\omega}^*$ on $\bm{\alpha}$, this gradient admits the following chain rule decomposition:
\begin{equation}\label{eq:4}
	\bm{g}_{\alpha} = \nabla_{\bm{\alpha}} \varphi(\bm{\omega}^*, \bm{\alpha}) + \nabla_{\bm{\alpha}} \bm{\omega}^*(\bm{\alpha})^{T} \nabla_{\bm{\omega}} \varphi(\bm{\omega}^*, \bm{\alpha}).
\end{equation}

Direct computation of the Jacobian $\nabla_{\bm{\alpha}} \bm{\omega}^*(\bm{\alpha})$ is prohibitively expensive. To circumvent this, we denote the lower-level objective as $f$. Using the first-order optimality condition $\nabla_{\bm{\omega}} f(\bm{\omega}^*, \bm{\alpha}) = 0$ and applying the implicit function theorem~\cite{The_Implicit_Function_Theorem}, we obtain:
\begin{equation}\label{eq:5}
	\nabla_{\bm{\alpha}} \bm{\omega}^*(\bm{\alpha}) = - \left[\nabla_{\bm{\omega} \bm{\omega}}^2 f(\bm{\omega}^*, \bm{\alpha})\right]^{-1} \nabla_{\bm{\omega} \bm{\alpha}}^2 f(\bm{\omega}^*, \bm{\alpha}).
\end{equation}
However, computing and inverting the Hessian matrix $\nabla_{\bm{\omega} \bm{\omega}}^2 f$ remains computationally intractable for large-scale models. We therefore adopt a rank-one outer-product approximation~\cite{Task_Oriented}, which yields:
\begin{equation}\label{eq:6}
	\nabla_{\bm{\omega} \bm{\omega}}^2 f \approx \nabla_{\bm{\omega}} f \, \nabla_{\bm{\omega}} f^T,
	\quad
	\nabla_{\bm{\omega} \bm{\alpha}}^2 f \approx \nabla_{\bm{\omega}} f \, \nabla_{\bm{\alpha}} f^T.
\end{equation}
This approximation is equivalent to a one-shot rank-one quasi-Newton update. Substituting Eqs.~\eqref{eq:5} and \eqref{eq:6} into Eq.~\eqref{eq:4}, we derive a computationally efficient hypergradient estimator that relies solely on first-order derivatives:
\begin{equation}\label{eq:7}
	\bm{g}_{\alpha} \approx \nabla_{\bm{\alpha}} \varphi - \frac{ \nabla_{\bm{\omega}} \varphi^{T} \nabla_{\bm{\omega}} f }{ \|\nabla_{\bm{\omega}} f\|^2 } \nabla_{\bm{\alpha}} f.
\end{equation}

Specifically, the implementation of BiLaLoRA is divided into two stages. In the Bilevel layer positioning stage, we solve the bilevel problem~\eqref{eq:3} to rank the importance of all candidate LoRA injection sites based on architectural parameters $\bm{\alpha}$. Subsequently, in the LoRA fine-tuning phase, we adapt the top-$k$ highest-ranked modules to achieve optimal performance on the target domain. The complete procedure is detailed in Algorithm~\ref{alg:lora}.

\subsection{Exploring Algorithmic Property}
\subsubsection{Cross-Model Flexibility}
To validate the model-agnostic property of BiLaLoRA, we applied it to four representative dehazing models (MSBDN, DeHamer~\cite{Dehamer}, ConvIR~\cite{ConvIR}, and DEA), all of which were uniformly pre-trained on the THaze~\cite{THaze} dataset. Fig.~\ref{fig:flexibilityquan} presents a comparative analysis of performance metrics before and after applying BiLaLoRA. Experiments demonstrate that BiLaLoRA effectively adapts to different network architectures while automatically positioning and optimizing performance bottleneck layers, substantially enhancing the performance of existing pre-trained models.

\begin{table}[t]
	\renewcommand\arraystretch{1.25} 
	\setlength{\tabcolsep}{0.5em}
	\centering
	\footnotesize
	\caption{\textbf{Quantitative results of full fine-tuning vs. BiLaLoRA}.}
	\vspace{-0.25cm}
	\begin{tabular}{|l|l|c|c|c|}
		\hline
		\multicolumn{2}{|l|}{Metric} & Fine-Tuning & BiLaLoRA & Rate\\ \hline
		\multirowcell{4}{\rotatebox[origin=c]{90}{\textit{\textbf{Performance}}}} 
		& FADE$\downarrow$      & 0.610 & 0.638 &$\downarrow$4.59\% \\ \cline{2-5}
		& BIQME$\uparrow$     & 0.617 & 0.611& $\downarrow$0.97\%\\ \cline{2-5}
		& Entropy$\uparrow$   & 7.569 & 7.572 &$\uparrow$0.04\%\\ \cline{2-5}
		& MUSIQ $\uparrow$   & 64.43 & 64.40& $\downarrow$0.05\%\\ \hline
		\multirowcell{4}{\rotatebox[origin=c]{90}{\textit{\textbf{Efficiency}}}}  
		& Train Time (H)    & 4.215 & 0.940 & $\downarrow$77.70\%\\ \cline{2-5}
	& Params. (M) & 3.653 & 3.764&$\uparrow$3.03\%\\ \cline{2-5}
		& FLOPs$^{\dagger}$  (G)  & 34.04 & 34.08&$\uparrow$1.18\%\\ \cline{2-5}
		& Runtime$^{\dagger}$(MS)& 3.702 & 3.735  &$\uparrow$0.89\%\\ \hline
	\end{tabular}
	\vspace{1mm} 
	\parbox{0.9\linewidth}{\scriptsize $^{\dagger}$FLOPs and Runtime are calculated on 256$\times$256 input.} 
	\vspace{-2em}
	\label{tab:comparison} 
\end{table}

\subsubsection{Cross-Domain Stability}
Subsequently, we employed DEA as the baseline model and conducted comprehensive experiments across four synthetic datasets (RIDCP~\cite{RIDCP}, ITS~\cite{reside}, OTS~\cite{reside}, and Haze4K~\cite{haze4k}). As demonstrated in Fig.~\ref{fig:stabilityquan}, BiLaLoRA consistently improved the performance of DEA models pre-trained on various datasets. Visual comparisons in Fig.~\ref{fig:stabilityqual} further substantiate that BiLaLoRA effectively recovers image content obscured by haze, leading to substantially enhanced dehazing performance in real-world applications.

\subsubsection{Full Fine-Tuning vs. BiLaLoRA}
As shown in Table~\ref{tab:comparison}, while full fine-tuning achieves satisfactory performance, BiLaLoRA attains comparable results with dramatically reduced training time by optimizing only adapter parameters. Notably, BiLaLoRA maintains similar FLOPs and parameters with negligible inference overhead. In addition, real-world dehazing tasks often encounter complex domain shifts. As illustrated in Fig.~\ref{fig:LoRA},  the fully fine-tuned model and adapter trained on daytime scenes cannot generalize to nighttime conditions. BiLaLoRA addresses this by training a separate nighttime adapter that achieves superior performance without costly full fine-tuning, thus facilitating efficient adaptation across diverse scenarios.

\begin{table*}[t]
	\setlength{\tabcolsep}{0.16em} 
	\renewcommand{\arraystretch}{1.35}
	\centering
	\scriptsize
	\caption{\textbf{Quantitative evaluations}. All metrics are computed on RTTS, URHI and Fattal.}
	\vspace{-1em}
	\begin{tabular}{|l|l|c|cccc|cccc|cccc|cccc|}
		\hline
		\multicolumn{3}{|c|}{Method}& \multicolumn{4}{c|}{{RTTS~\cite{reside}}} & \multicolumn{4}{c|}{{URHI~\cite{reside}}} & \multicolumn{4}{c|}{{Fattal~\cite{fattal}}} & \multicolumn{4}{c|}{{Average}} \\
		\hline
		\multicolumn{2}{|c|}{Name}&Venue& FADE & BIQME& Entropy & MUSIQ & FADE & BIQME& Entropy& MUSIQ& FADE & BIQME & Entropy& MUSIQ &FADE & BIQME  & Entropy & MUSIQ\\
		\hline
		\multirow{4}{*}{\rotatebox{90}{Synthetic}}&MSBDN~\cite{MSBDN} &CVPR 20&1.483&0.549&7.273&52.93&1.517&0.542&7.264&54.68&0.613&0.555&7.408&63.71&1.204&0.568&7.315&57.11\\
		&DeHamer~\cite{Dehamer}  &CVPR 22&1.806&0.542&7.215&52.90&1.853&0.537&7.217&54.67&0.756&0.552&7.411&64.31&1.472&0.544&7.281&57.29\\
		&C$^{2}$PNet~\cite{C2PNet}  &CVPR 23&2.050&0.531&7.168&54.18&2.054&0.524&7.157&56.48&0.720&0.551&7.399&65.00&1.608&0.535&7.241&58.55\\
		&DEA~\cite{DEA}  &TIP 24&1.781&0.541&7.196&53.13&1.891&0.534&7.198&54.89&0.696&0.557&7.423&64.23&1.456&0.544&7.272&57.42\\
		\hline
		\multirow{4}{*}{\rotatebox{90}{All-In-One$\;$}}&PromptIR~\cite{PromptIR} &NeurIPS 23&1.765&0.546&7.189&53.80&1.747&0.544&7.390&55.97&0.668&0.558&7.498&64.96&1.393&0.549&7.359&58.24\\
		&DiffUIR~\cite{DiffuIR}  &CVPR 24&2.132&0.531&7.172&54.81&2.014&0.527&7.181&56.39&0.871&0.537&7.373&64.95&1.672&0.532&7.242&58.72\\
		&MoCE-IR~\cite{MoCEIR}  &CVPR 25&1.922&0.539&7.191&54.21&1.664&0.541&7.217&57.25&0.678&0.557&7.409&65.01&1.421&0.546&7.272&58.82\\
		&FoundIR~\cite{FoundIR}  &ICCV 25&1.760&0.553&7.275&54.96&1.762&0.550&7.301&57.12&0.759&0.560&7.376&65.83&1.427&0.554&7.317&59.30\\
		\hline
		\multirow{8}{*}{\rotatebox{90}{Real}}&DAD~\cite{DAD}  &CVPR 20& 1.131 & 0.561 & 7.413 & 49.34 & 1.099 & 0.566 & 7.439 & 50.83 & 0.487 & 0.589 & 7.487 & 59.38 &0.905&0.572&7.446&53.18\\
		&PSD~\cite{PSD}  &CVPR 21& 1.143 & 0.524 & 7.276 & 52.81 & 0.937 & 0.517 & 7.252 & 55.99 & 0.438 & 0.554 & 7.463 & 63.80&0.839&0.532&7.330&57.53 \\
		&D4~\cite{D4}  &CVPR 22& 1.404 & 0.556 & 7.179 & 53.57 & 1.116 & 0.549 & 7.236 & 56.27 & 0.457 & 0.537 & 7.372 & 64.14&0.992&0.547&7.262&57.99 \\
		&RIDCP~\cite{RIDCP}  &CVPR 23& 0.955 & \underline{0.600} & 7.541 & 59.14 & 0.922 & \underline{0.603} & 7.559 & 61.73 & 0.396 & 0.604 & 7.468 & 66.10 &0.758&\underline{0.602}&7.523&62.32\\
		&KANet~\cite{KANet}  &TPAMI 24& 0.870 & 0.583 & 7.517 & 54.54 & \textbf{0.867} & 0.589 & 7.555 & 56.75 & 0.338 & 0.560 & 7.527 & 65.35 &0.692&0.577&7.533&58.88\\
		&CoA~\cite{CoA} &CVPR 25&0.859&0.593&\textbf{7.579}&53.43&0.927&0.596&\underline{7.592}&55.93&\underline{0.314}&\textbf{0.618}&\textbf{7.585}&63.38&0.700&\underline{0.602}&\textbf{7.585}&57.58\\
		&IPC~\cite{IPC}  &CVPR 25& 1.105 & 0.592 & 7.469 & \underline{59.61} & 1.103 & 0.592 & 7.512 & \underline{62.22} & 0.368 & 0.593 & 7.471 & \underline{67.58}&0.858&0.592&7.484&\underline{63.14} \\
		&PHATNet~\cite{PHATNet} &ICCV 25&\underline{0.845}&0.585&7.349&56.43&0.892&0.582&7.390&58.34&0.331&0.604&7.498&66.87&\underline{0.689}&0.590&7.412&60.55\\
		\hline
		\multicolumn{2}{|c|}{BiLaLoRA}  &Ours& \textbf{0.752} & \textbf{0.611} & \underline{7.576} & \textbf{61.77} & \underline{0.881} & \textbf{0.615} & \textbf{7.599} & \textbf{63.52} & \textbf{0.281} & \underline{0.607} & \underline{7.541} & \textbf{67.92} &\textbf{0.638}&\textbf{0.611}&\underline{7.572}&\textbf{64.40}\\
		\hline
	\end{tabular}
	\label{tab:multidataset}
	\vspace{-0.15cm}  
\end{table*}

\begin{figure*}[t]
	\centering
	\includegraphics[width=0.98\textwidth]{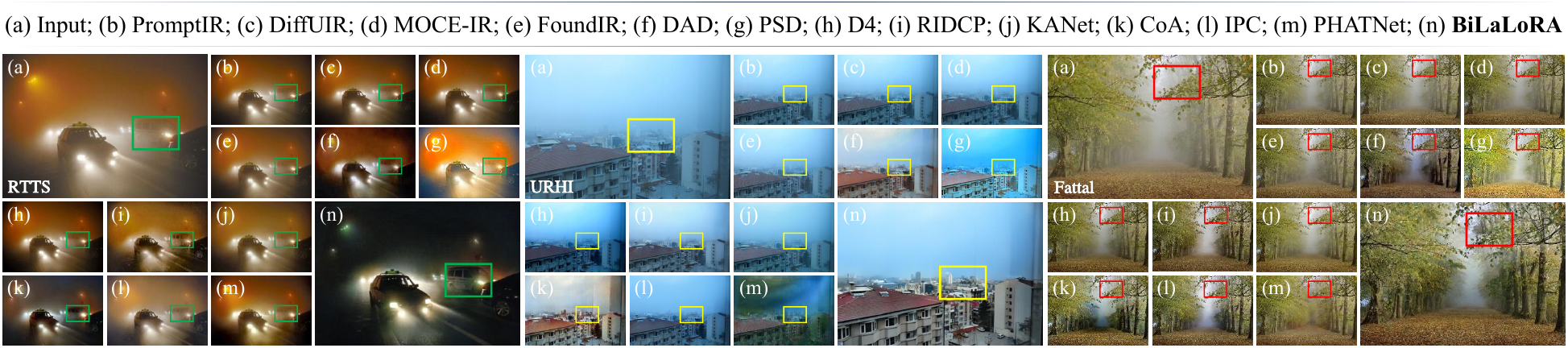}
	\vspace{-0.2cm}
	\caption{\textbf{Visual comparisons on different datasets}. Boxes indicate specific areas that highlight differences.}
	\label{fig:comparisons}
	\vspace{-1em}
\end{figure*}

\section{Experimental Results}

\subsection{Implementation Details}

\textbf{Training settings.}
The BiLaLoRA was implemented using the PyTorch framework on a single NVIDIA 4090 GPU. For all experiments, we adopted DEA as the baseline and employed the Adam optimizer to update parameters, with $\beta_1$, $\beta_2$, and $\varepsilon$ set to 0.9, 0.999, and $1 \times 10^{-8}$, respectively. During the pre-training stage, the baseline model was trained on the THaze dataset using $\ell_1$ loss, with the learning rate initialized at $1 \times 10^{-4}$ and gradually decayed to $1 \times 10^{-6}$ using cosine annealing scheduling. For domain adaptation in the BiLaLoRA stage, two domain-specific adapters were developed for daytime and nighttime dehazing. These were trained respectively using 500 real daytime haze images~\cite{fade} and 100 nighttime images from NHRW~\cite{NHRW}, with both datasets being split equally into training and validation sets. We conducted bilevel layer-positioning and LoRA fine-tuning on the top-$3$ layers using the H2C loss, maintaining a learning rate of $1 \times 10^{-6}$. For the LoRA modules, we set the scaling factor to $\gamma = 2$, with the rank configured to $r = 8$. During all training stages, we augmented the training data by cropping random $256 \times 256$ patches from the images, which were then subjected to random 90°, 180°, and 270° rotations and horizontal flipping.

\noindent\textbf{Benchmarks and metrics.}
To thoroughly evaluate the model, we conducted experiments on three real datasets: RTTS, URHI, and Fattal, which comprise 4,322, 150, and 31 images, respectively. Additionally, we assessed the model's generalization capability on the HazyDet~\cite{hazydet}, Dense-Haze~\cite{densehaze}, and O-Haze~\cite{O-HAZE} datasets. For quantitative assessment, we employed four no-reference metrics: the fog density assessment method (FADE)~\cite{fade}, the blind image quality metric for enhanced images (BIQME)~\cite{biqme}, the image entropy assessment index (Entropy)~\cite{entropy}, and the multi-scale image quality transformer (MUSIQ)~\cite{musiq}.

\begin{figure*}[t]
	\centering
	\includegraphics[width=0.995\textwidth]{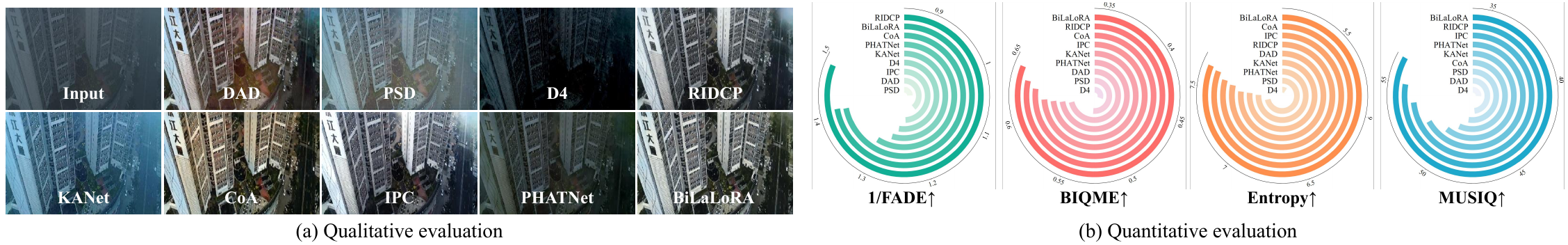}
	\vspace{-0.3cm}
	\caption{\textbf{Generalization evaluation on HazyDet dataset}. All models were evaluated on the testing dataset without retraining.}
	\label{fig:generalizationa}
	\vspace{-1em} 
\end{figure*}

\begin{figure}[t]
	\centering
	\includegraphics[width=0.47\textwidth]{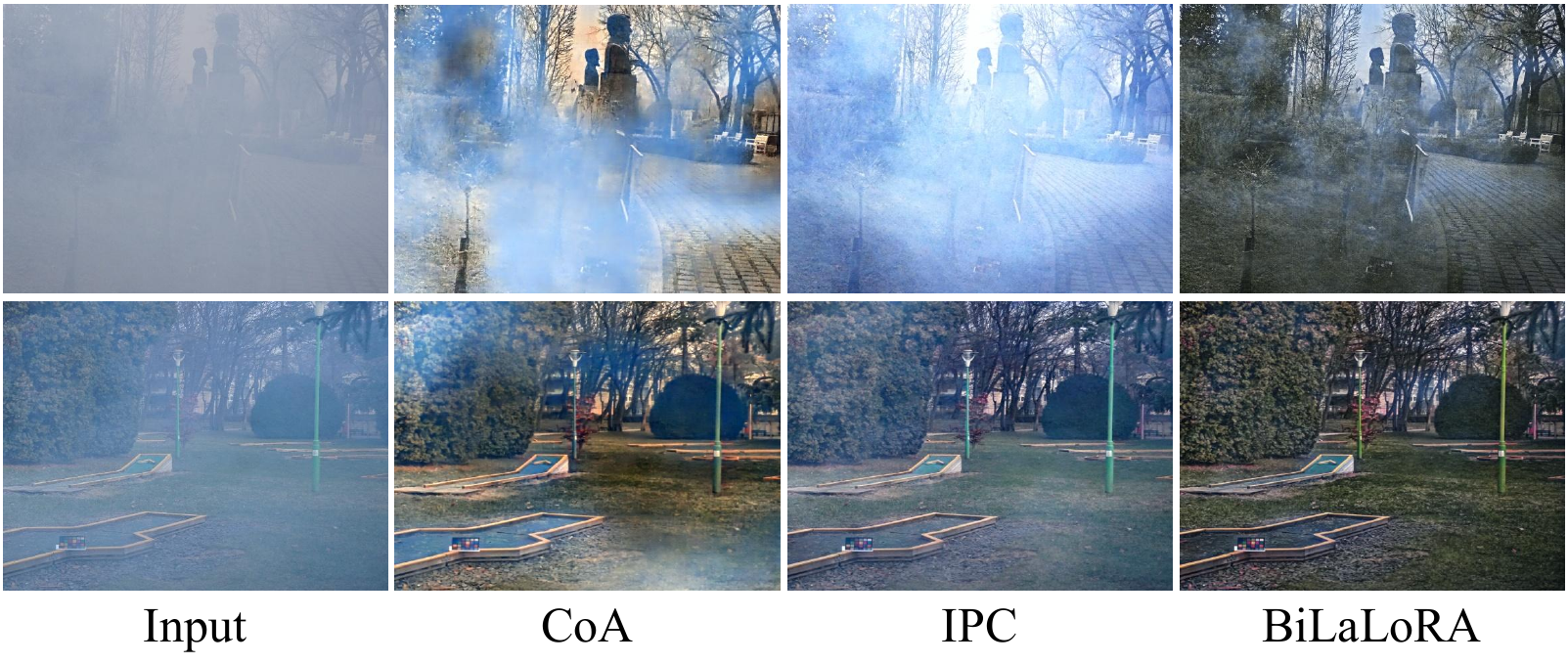}
	\vspace{-0.3cm}
	\caption{\textbf{Generalization evaluation on other real datasets}. Dense-Haze (Top) and O-Haze (Bottom). All models were evaluated on the testing dataset without retraining.}
	\label{fig:generalizationb}
	\vspace{-1em} 
\end{figure}

\subsection{Performance Evaluation}

To comprehensively evaluate the performance of BiLaLoRA, we conducted extensive quantitative and qualitative analyses, comparing it against various state-of-the-art specialized dehazing methods and general all-in-one image restoration models. Additional experimental details and analytical results are provided in the supplementary material.

\noindent\textbf{Quantitative Evaluation.} 
The quantitative results in Table~\ref{tab:multidataset} demonstrate the strong performance of BiLaLoRA. Compared with state-of-the-art methods, our model ranks first or second across key evaluation metrics, fully demonstrating its outstanding performance in real image dehazing.

\noindent\textbf{Qualitative Evaluation.} 
Fig.~\ref{fig:comparisons} presents qualitative comparisons on the RTTS, Fattal, and URHI datasets. All-in-one models demonstrate limited generalization to real-world scenes. Specialized dehazing methods exhibit various limitations: DAD, D4, and KANet fail to adequately handle colored haze; PSD suffers from overexposure and color shift; and PHATNet produces visual artifacts. While RIDCP, CoA, and IPC yield comparatively better results, they still lack in detail fidelity and naturalness. In contrast, BiLaLoRA effectively removes haze while better preserving fine details and maintaining natural appearance.

\subsection{Generalization Evaluation}
To further validate the robustness of BiLaLoRA, we conducted additional evaluations under more challenging conditions. As illustrated in Fig.~\ref{fig:generalizationa}, when evaluated on the real-world UAV-perspective haze dataset HazyDet, our method successfully recovers scene details obscured by haze while effectively preventing color distortion, with qualitative results confirming its efficacy. Moreover, as shown in Fig.~\ref{fig:generalizationb}, the performance of previously competitive methods such as CoA and IPC deteriorates substantially on dense haze datasets including Dense-Haze and O-Haze, where they almost fail to generate meaningful dehazing outputs. By comparison, BiLaLoRA maintains consistent performance even under these extreme conditions, demonstrating excellent generalization capability across diverse scenarios.

\begin{table}[t]
	\renewcommand\arraystretch{1.2} 
	\setlength{\tabcolsep}{0.04em}
	\centering
	\scriptsize
	\caption{\textbf{Quantitative evaluation of ablation study}.}
	\vspace{-0.3cm}
	\begin{tabular}{|c|cc|cc|cccc|}
		\hline
		\multirow{2}{*}{Model}&\multicolumn{2}{c|}{{H2C Loss}}&\multicolumn{2}{c|}{{Layer-Positioning}}&\multicolumn{4}{c|}{{Averaged Performance}}\\
		\cline{2-9}
		&Positive&Negative&$\;$Naïve$\;$&Bilevel&$\;$FADE$\;$ &$\;$BIQME$\;$&$\;$Entropy$\;$&$\;$MUSIQ$\;$\\
		\hline
		$\text{Q}_{\mathtt{a}}$&$\times$&$\times$&$\times$&$\times$& 1.018 & 0.582 & 7.438 & 62.05\\
		\hline
		$\text{Q}_{\mathtt{b}}$&\checkmark&$\times$&$\times$&$\times$&0.862&0.589&7.544&62.23\\
		$\text{Q}_{\mathtt{c}}$&\checkmark&$\times$&\checkmark&$\times$&0.705&0.592&7.559&62.35\\
		$\text{Q}_{\mathtt{d}}$&\checkmark&$\times$&$\times$&\checkmark&0.680&0.601&7.563&62.57\\
		\hline
		$\text{Q}_{\mathtt{e}}$&$\times$&\checkmark&$\times$&$\times$&0.774&0.561&7.533&60.47\\
		$\text{Q}_{\mathtt{f}}$&$\times$&\checkmark&\checkmark&$\times$&0.745&0.579&7.537&61.04\\
		$\text{Q}_{\mathtt{g}}$&$\times$&\checkmark&$\times$&\checkmark&0.712&0.584&7.541&61.23\\
		\hline
		$\text{Q}_{\mathtt{h}}$&\checkmark&\checkmark&$\times$&$\times$&0.774&0.600&7.559&63.31\\
		$\text{Q}_{\mathtt{i}}$&\checkmark&\checkmark&\checkmark&$\times$&0.662&0.607&7.566&64.07\\
		\hline
		Ours&\checkmark&\checkmark&$\times$&\checkmark& \textbf{0.638} & \textbf{0.611} & \textbf{7.572}& \textbf{64.40}\\
		\hline
	\end{tabular}\label{tab:ablation}
	\vspace{-1.4em} 
\end{table}

\section{Algorithmic Analyses}

\subsection{Effects of Text-Directed Loss}
\begin{figure*}[t]
	\centering
	\includegraphics[width=0.97\textwidth]{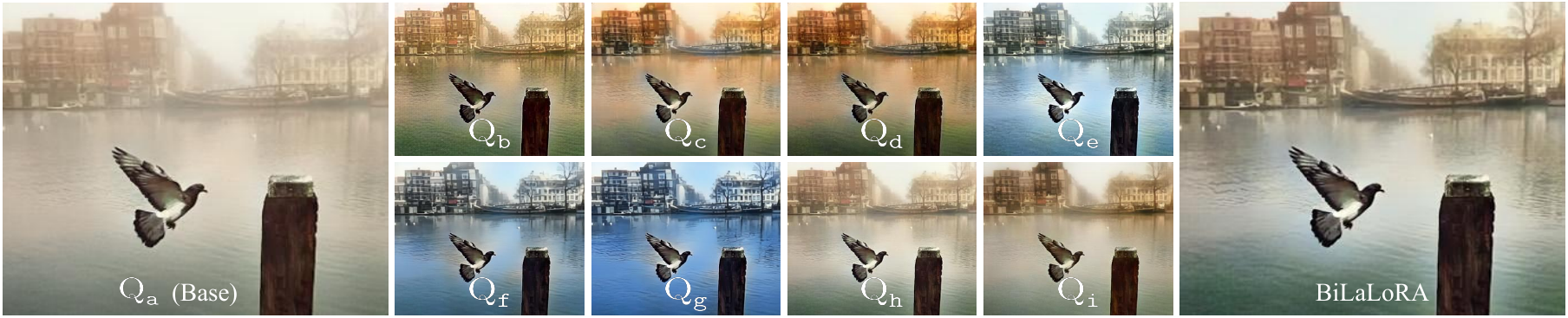}
	\vspace{-0.1cm}
	\caption{\textbf{Qualitative results of ablation study}. Visual comparison of different combinations in Table~\ref{tab:ablation}, with data from URHI.}
	\label{fig:ablation}
	\vspace{-1.2em} 
\end{figure*}

To validate the efficacy of the directional guidance mechanism within the H2C loss, we conduct an ablation study comparing the full loss against two degraded variants, as presented in Table~\ref{tab:ablation}. After removing the negative guidance, the model aligns the output image features solely with the semantic representation of the positive text. This simplified configuration causes the H2C loss to drive the outputs toward a singular positive semantic target, thereby neglecting content consistency with the input.  As illustrated in Fig.~\ref{fig:ablation}, while this variant achieves partial haze removal, it introduces substantial color distortion artifacts. Conversely, in the absence of positive text guidance, the optimization objective becomes dominated by excessive suppression of haze related features, ultimately culminating in over-dehazing phenomena. These findings demonstrate that the H2C loss, through the synergistic interplay of positive and negative textual constraints, establishes a well-defined semantic optimization trajectory for the dehazing process, ensuring effective haze removal while maximally preserving the structural integrity of the original scene.

\begin{figure}[t]
	\centering
	\includegraphics[width=0.47\textwidth]{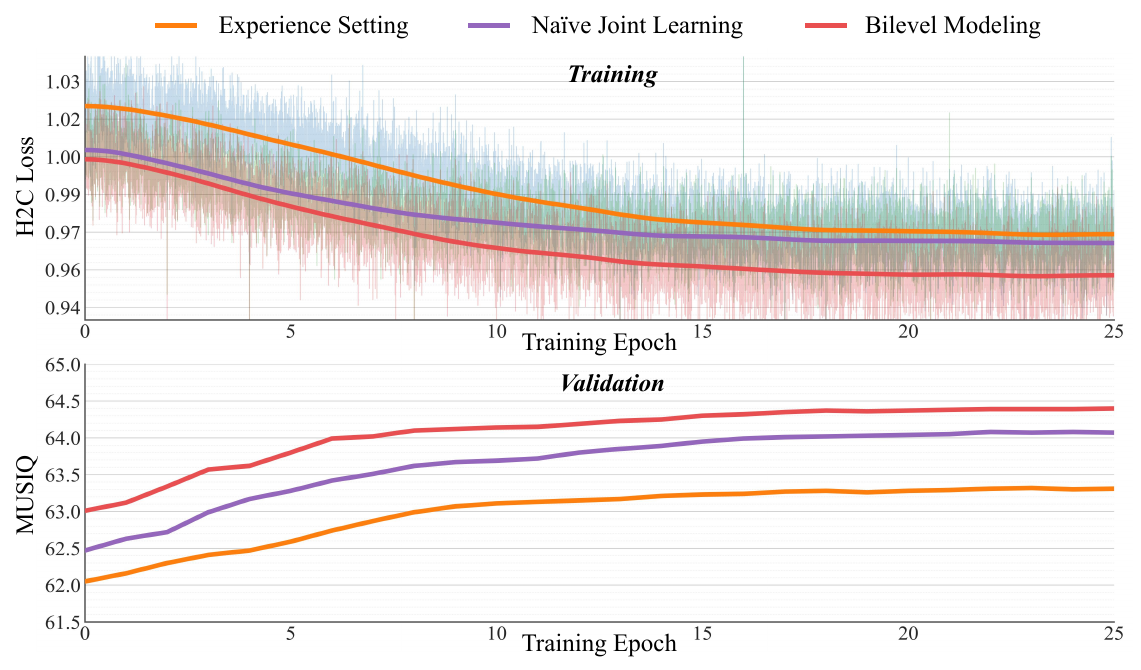}
	\vspace{-0.2cm}
	\caption{\textbf{Convergence behaviors}. H2C loss convergence (Top) and MUSIQ improvement (Bottom) over training epochs. }
	\label{fig:convergence}
	\vspace{-1em} 
\end{figure}
\subsection{Necessity of Bilevel Modeling}
Furthermore, we compared the bilevel modeling with the experience setting and naïve joint learning paradigms. The experience setting relies on heuristic manual selection of adaptation layers, a methodology that cannot guarantee adaptability across different architectures and domains. Joint optimization improves upon this by introducing learnable architecture parameters. However, simultaneously optimizing both architecture parameters and weight increments on the same training set causes the architecture search to overly rely on loss feedback from the training set. In contrast, bilevel optimization decouples these objectives by updating architecture parameters based on the validation set, thereby directly aligning the architecture search with generalization performance. As indicated in Table~\ref{tab:ablation}, the bilevel modeling strategy yields substantial performance gains over both the experience setting and naïve joint learning.

Moreover, the convergence analysis presented in Fig.~\ref{fig:convergence} reveals that manual layer selection is inherently constrained by its predetermined choices, often leading to suboptimal adaptation and limited flexibility. In contrast, bilevel modeling exhibits significantly more stable convergence dynamics and sustained performance improvements relative to joint optimization, demonstrating its ability to automatically pinpoint and fine-tune bottleneck layers.
\begin{figure}[t]
	\centering
	\includegraphics[width=0.47\textwidth]{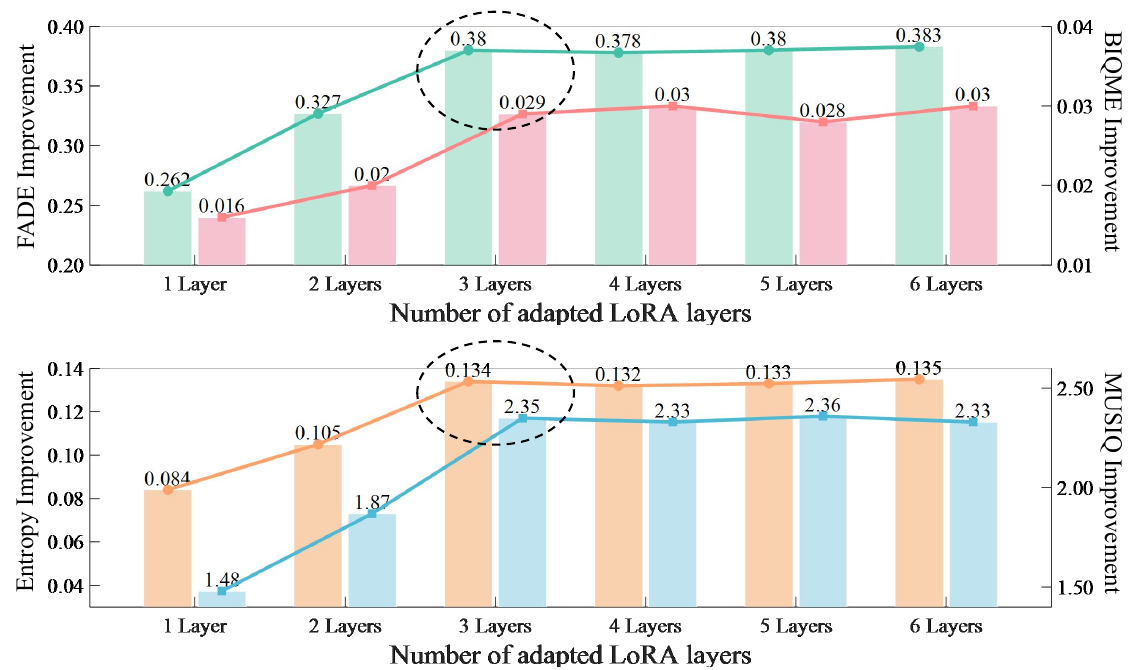}
	\vspace{-0.2cm}
	\caption{\textbf{Parameter analysis}. The black dashed circle indicates the optimal selection. }
	\label{fig:parameters}
	\vspace{-1em} 
\end{figure}
\subsection{Influence of Layer Number}
The quantity of adapter layers is a critical factor of a model's adaptation capability for the target domain and its parameter efficiency. To explore this impact, we assessed the comprehensive performance of BiLaLoRA under different LoRA layer activation states. As shown in Fig.~\ref{fig:parameters}, model performance improves steadily as the number of adaptation layers increases, with performance peaking at three layers. However, beyond this point, the performance improvement curve flattens, demonstrating clear diminishing marginal returns. These results indicate that additional adapter layers not only fail to deliver significant performance improvements but also lead to parameter redundancy and unnecessary computational overhead.

\section{Concluding Remarks}
BiLaLoRA automatically pinpoints and optimizes the performance bottleneck layers, significantly improving cross-domain performance with minimal parameter overhead. Leveraging the inherent plug-and-play nature of LoRA, BiLaLoRA provides a highly flexible and effective domain adaptation solution for real image dehazing.

Our future work will explore the application of BiLaLoRA to other low-level vision tasks in diverse real-world scenarios. Additionally, we plan to investigate more refined semantic guidance mechanisms and cross-domain adaptation techniques to address restoration challenges under severe degradation conditions.

\section*{Acknowledgments}
This research was supported by the foundations of Guangdong Basic and Applied Basic Research Foundation (No.~2024A1515011563), Natural Science Foundation of Hangzhou (No.~2025SZRJJ1901), National Natural Science Foundation of China (No.~62506060).